\documentclass{article}
\PassOptionsToPackage{table}{xcolor}
\usepackage{iclr2026_conference,times}

\usepackage{amsmath,amsfonts,bm}

\def\eqref#1{equation~\ref{#1}}

\def\1{\bm{1}}

\DeclareMathAlphabet{\mathsfit}{\encodingdefault}{\sfdefault}{m}{sl}
\SetMathAlphabet{\mathsfit}{bold}{\encodingdefault}{\sfdefault}{bx}{n}

\usepackage{hyperref}
\usepackage{url}
\usepackage{graphicx}
\usepackage{amsmath,amssymb}
\usepackage{xcolor}
\usepackage{float}
\usepackage{wrapfig}
\usepackage{needspace}

\graphicspath{{figs/}}

\newcommand{\figurelayout}{\centering\setlength{\parskip}{0pt}}
\newcommand{\wrapfigureraise}{\vspace*{-30.5pt}}
\newcommand{\figurecaption}{\par\nointerlineskip\caption}

\definecolor{actionblue}{HTML}{2E6688}
\definecolor{noiseochre}{HTML}{95651D}
\newcommand{\actionterm}[1]{{\color{actionblue}#1}}
\newcommand{\noiseterm}[1]{{\color{noiseochre}#1}}

\title{EigenDEXplore: Structured Exploration for Dexterous Manipulation with Human Priors}
\newcommand{\ours}{EigenDEXplore}

\author{Harsh Gupta\thanks{Equal contribution.} \quad Tyler Ga Wei Lum\footnotemark[1] \quad Changhao Wang \quad Chuer Pan \\
\textbf{C. Karen Liu \quad Jeannette Bohg \quad Shuran Song} \\
Stanford University}

\iclrfinalcopy
\begin{document}
\maketitle
\ificlrfinal\lhead{Preprint}\vspace*{-26pt}\fi
\begin{abstract}
Dexterous manipulation poses a challenging high-dimensional optimization problem, as useful behaviors require coordinated motion across many hand joints. In reinforcement learning (RL) and sampling-based trajectory optimization, exploration commonly relies on independent robot joint perturbations, making coordinated behaviors difficult to discover. Prior work reduces this search space for grasp learning using low-dimensional spaces of coordinated joint motions learned from human hand data, but this restricts the expressivity required for general manipulation. Some combine learned and joint-space actions to restore expressivity, but this increases dimensionality and introduces redundancy. We study these effects across diverse manipulation settings, varying action dimensionality, exploration strategy, and the source of human data. Our experiments suggest that human-motion priors are most effective when used to structure exploration rather than change the action representation. Motivated by this finding, we propose \ours{}, which induces correlated exploration by adding perturbations along human-derived eigenvectors to independent joint-space noise, leaving the action space unchanged. Across multiple dexterous hands, \ours{} consistently outperforms joint-space and learned action-space baselines in grasping, in-hand reorientation, and contact-rich manipulation. These gains span unstructured and reference-guided RL, trajectory optimization, and sim-to-real deployment, and are largest in settings with less reward shaping and curriculum design. Project page: \ificlrfinal\href{http://eigendexplore.github.io/}{\nolinkurl{eigendexplore.github.io}}\else\href{https://sites.google.com/view/eigendexplore}{\nolinkurl{sites.google.com/view/eigendexplore}}\fi.
\end{abstract}

\begingroup
\setlength{\intextsep}{4pt}
\begin{figure}[H]
\figurelayout
\includegraphics[width=\linewidth]{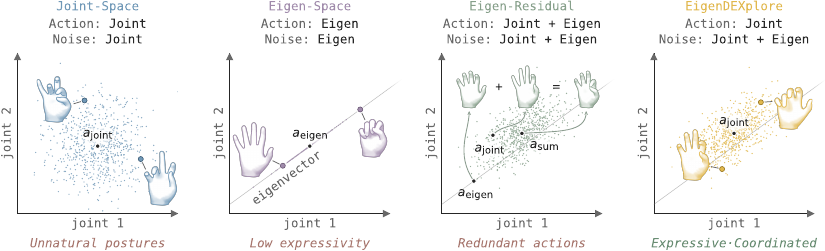}
\figurecaption{
\textbf{Action spaces and exploration noise.} Joint-Space perturbs joints independently, which can produce unnatural hand configurations. Eigen-Space encourages coordinated motion but constrains actions to a learned PCA subspace. Eigen-Residual restores full joint-space expressivity by combining joint and eigen actions, but it introduces redundancy in the action parameterization. \ours{} combines joint and eigen noise while retaining a joint-space action representation.}
\label{fig:exploration-teaser}
\end{figure}
\endgroup

\section{Introduction}

Dexterous hands enable versatile grasping, in-hand manipulation, and precise interaction with tools and articulated objects. Discovering these contact-rich behaviors, however, is challenging because effective manipulation relies on highly coordinated finger motions that occupy only a small fraction of the hand’s high-dimensional action space. Reinforcement learning (RL) and sampling-based trajectory optimization commonly explore this space using independent Gaussian perturbations across joints, making it difficult to discover such coordinated behaviors efficiently. As a result, these methods often require substantial computation and carefully engineered rewards, costs, or curricula.

Learning from human hand motion offers a way to capture useful coordination patterns, providing guidance that would otherwise need to be designed for each task. Prior work captures these patterns in low-dimensional action spaces, including linear eigen-spaces learned using PCA~\citep{eigengrasps2007,dexfunc} and nonlinear latent spaces learned using VAEs~\citep{dimou2021synergies,taskorientedgrasping}. These spaces let the policy directly command coordinated movements such as grasping and twisting (see Fig.~\ref{fig:action-bases}). Sampling actions corresponding to these modes during RL exploration can also provide more informative feedback per sample for policy updates and accelerate learning~\citep{dexfunc,dextrahg}. However, these reduced action spaces cannot express the independent joint control required for general manipulation~\citep{grasp2act}. Another approach combines these learned and independent joint-space actions to restore expressivity~\citep{lucid}, but increases the action dimensionality and introduces redundancy. It remains unclear whether the benefits come from directly controlling coordinated motions, the noise-induced exploration, or changes to action-space dimensionality and expressivity.

To disentangle these effects, we systematically compare action representations and exploration distributions across diverse manipulation tasks and robot hands. \textit{These comparisons suggest that the primary benefit of human-motion priors comes from correlated exploration, while action constraints or redundancy can hinder performance.} Joint-space actions already allow the policy to express both coordinated and independent joint movements. The difficulty is discovering useful joint combinations, which correlated noise makes easier to sample. Motivated by this finding, we introduce \textbf{\ours{}}, a simple method that incorporates this human-motion prior into the sampling distribution while retaining the original joint-space actions. It combines two noise components: perturbations along the learned PCA directions encourage coordinated motion, while independent joint-space noise maintains exploration coverage. The prior is derived from hand motion recovered from large-scale human video data~\citep{egosuite}, without requiring task-specific trajectories. After RL training, the eigenbasis can be discarded and the policy deployed deterministically.

We compare \ours{} with joint-space and learned eigen-space baselines across unstructured RL with SimToolReal~\citep{simtoolreal} and DeXtreme~\citep{dextreme}, reference-guided bimanual RL with DexMachina~\citep{dexmachina}, and trajectory optimization with SPIDER~\citep{spider}, spanning six dexterous hands. \ours{} consistently outperforms them, with 34\% more consecutive successes in SimToolReal at a fixed training budget and 40\% more in DeXtreme. These gains transfer to hardware, raising SimToolReal average progress across 4 tasks from 33\% to 82\% and DeXtreme consecutive successes by 41\%. In DexMachina, it improves average object-tracking accuracy by 28\% without contact reward shaping or curricula, with larger gains on higher-DoF hands, enabling learning where joint-space sampling struggles or fails. In SPIDER, it reduces retargeting cost by 16\% on average across six hands, again most on higher-DoF hands. We also find that large-scale egocentric video~\citep{egosuite} yields stronger human-motion priors than smaller, task-focused motion-capture datasets~\citep{grab,arctic}, and validate our choice of retained PCA dimensionality through exploration-distribution experiments.

In summary, our main contribution is \ours{}, a method that guides exploration with human-derived coordination patterns while preserving joint-space actions. We identify correlated exploration as the primary source of improvement and demonstrate the method’s effectiveness across reinforcement learning, trajectory optimization, and sim-to-real deployment. We release benchmarking code, a plug-and-play implementation, and pretrained PCA models for six dexterous hands.

\section{Related Work}
\label{sec:related-work}

\textbf{Human priors for manipulation.} Human data can guide dexterous manipulation by providing examples of useful behavior and desired outcomes. Demonstrations and pose priors support learning through policy supervision~\citep{dapg,dexmv,guzey2025dexterity}, initialization~\citep{demostart,pgdm}, and reward design~\citep{dexvip}. Hand and object trajectories provide task objectives for RL~\citep{human2sim2robot,grasp2act,dexmachina,dexplore,maniptrans,dextrack,dexh2r,vividex} and trajectory optimization~\citep{spider,doasido,quasisim}, allowing robot motions to adapt to embodiment differences and contact dynamics. Procedurally trained controllers can also execute human-derived goals at deployment~\citep{dexgen,simtoolreal,lucid,dex4d}. The success of task guidance from even a single video or demonstration motivates examining how larger human datasets can support learning across tasks.

\textbf{Action representations for dexterous manipulation.} Early studies showed that a few coordinated patterns capture much of the variation in human grasp postures~\citep{santello1998}, motivating eigen-space grasp planning~\citep{eigengrasps2007,ciocarlie2009}. Encoding joint coordination in control representations supports RL and transfer in dexterous manipulation~\citep{dexfunc,dextrahg,taskorientedgrasping,lamp,crossdex,pchands,berg2024sar} and lowers trajectory-optimization cost for a fixed number of sampled trajectories~\citep{vqace}. However, low-dimensional eigen-space control can restrict dexterity~\citep{dextrahg,adept}, including the fine adjustments needed to establish sufficient contact during tool manipulation~\citep{grasp2act}. Adding eigen-space residuals to joint-space actions preserves full expressivity and can improve learning over joint-only control despite introducing redundant action coordinates~\citep{lucid}. Reduced dimensionality therefore cannot explain all reported gains, raising the question of how these interfaces shape exploration.

\textbf{Structure in exploration.} These interfaces can turn independent action noise into correlated joint perturbations, confounding the benefits of coordinated commands and coordinated sampling. Temporal correlations~\citep{gsde,pinknoise,coloredppo} and policy-derived correlations across actuators~\citep{lattice} show how structuring exploration noise can improve learning efficiency. Human motion offers such structure, with the source data and learned representation determining which movements are encouraged. Existing priors often use grasp-focused recordings or small task-specific demonstration sets, with learned representations that vary in implementation and dimensionality~\citep{dextrahg,taskorientedgrasping,crossdex,lamp}. Our study separates the effects of human-motion priors on exploration from their use in the action representation, and examines how source data and prior dimensionality influence manipulation performance.

\section{Method}
\label{sec:method}

\begin{figure}[t]
\figurelayout
\includegraphics[width=\linewidth,trim=0 1.8bp 0 3.5bp,clip]{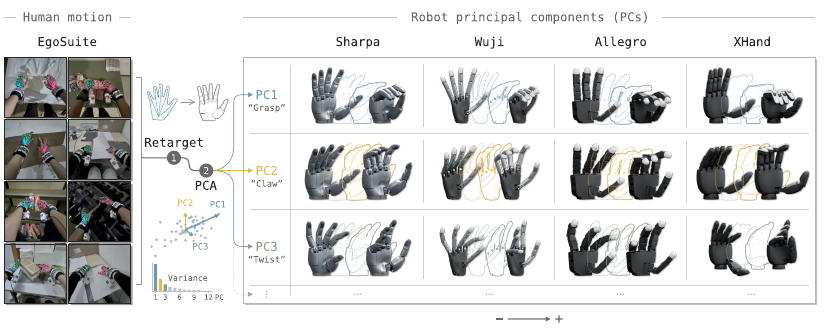}
\figurecaption{
\textbf{Constructing robot principal components from human motion.} We retarget EgoSuite human hand motion to each robot hand and apply PCA in its joint space. We visualize the first three principal components for Sharpa, Wuji, Allegro, and XHand.}
\label{fig:action-bases}
\end{figure}

\ours{} uses human hand coordination to guide exploration while preserving native joint-space control. We retarget human motion to each robot hand and use PCA to extract a reusable basis of coordinated joint movements (Sec.~\ref{sec:coordination-prior}). We then combine perturbations along these directions with independent joint-space noise (Sec.~\ref{sec:coordinated-exploration}), sampling around the policy mean in reinforcement learning (Sec.~\ref{sec:reinforcement-learning}) and the nominal control sequence in trajectory optimization (Sec.~\ref{sec:method-integration}).

\subsection{Human-Derived Coordination Prior}
\label{sec:coordination-prior}

We use a dataset of human hand postures to construct a coordination prior for each robot hand. We pool postures from both hands by mirroring one hand into the other's coordinate frame, forming a shared canonical representation. We adapt the kinematic retargeting procedure from~\citep{maniptrans} to map human hand motion to each robot, with hand-specific alignment, scaling, and finger correspondences. We apply PCA to the retargeted joint angles to identify common patterns of joint coordination, with eigenvalue $\lambda_i$ measuring the variance captured by component $i$. To retain a chosen fraction of the human dataset variance, we select the first $k$ principal components needed to reach that threshold. The resulting $k$ depends on both the hand and dataset. For example, retaining 90\% of the variance gives $k=9$ for Sharpa (22 joints) and $k=6$ for XHand (12 joints). We express the first $k$ directions in the policy's normalized joint-action coordinates (each joint scaled to $[-1,1]$ by its limits) as the rows of the eigenbasis $E\in\mathbb{R}^{k\times d}$, where $d$ is the number of hand joints. Fig.~\ref{fig:action-bases} illustrates the leading directions across hands.

\subsection{Action Spaces and Exploration Noise}
\label{sec:coordinated-exploration}

We compare \ours{} with three action formulations (Fig.~\ref{fig:exploration-teaser}): Joint-Space, which does not use $E$, as well as Eigen-Space and Eigen-Residual, which build $E$ into the action representation. These comparisons help disentangle the effect of incorporating human coordination into the action representation versus using it only to structure exploration. They also characterize the associated tradeoffs in action-space dimensionality, expressivity, and exploration structure (Sec.~\ref{sec:action-space-comparison}). Let $a_q\in\mathbb{R}^d$ and $a_e\in\mathbb{R}^k$ denote the joint- and eigen-space coordinates of a command before exploration. Let $q^\star\in\mathbb{R}^d$ denote the sampled absolute joint-position target in these normalized coordinates.
\begin{equation}
    \begin{alignedat}{3}
    &\text{Joint-Space}\ (q) &\quad&q^\star = \actionterm{a_q}+\noiseterm{\sigma_q\odot\eta_q}
    &\quad&\text{action dim}=d\\
    &\text{Eigen-Space}\ (e) &&q^\star = E^\top\!\left(\actionterm{a_e}+\noiseterm{\sigma_e\odot\eta_e}\right)
    &&\text{action dim}=k\\
    &\text{Eigen-Residual}\  &&q^\star = \actionterm{a_q}+\noiseterm{\sigma_q\odot\eta_q}
    +E^\top\!\left(\actionterm{a_e}+\noiseterm{\sigma_e\odot\eta_e}\right)
    &&\text{action dim}=d+k\\
    &\text{\ours{}}\  &&q^\star = \actionterm{a_q}+\noiseterm{\sigma_q\odot\eta_q}
    +E^\top(\noiseterm{\sigma_e\odot\eta_e})
    &&\text{action dim}=d
    \end{alignedat}
    \label{eq:formulations}
\end{equation}
Here, $\eta_q\sim\mathcal{N}(0,I_d)$ and $\eta_e\sim\mathcal{N}(0,I_k)$ are independent standard Gaussian vectors, $\sigma_q\in\mathbb{R}^d$ and $\sigma_e\in\mathbb{R}^k$ are perturbation standard deviations, and $\odot$ denotes elementwise multiplication. We initialize $\sigma_{e,i}\propto\sqrt{\lambda_i}$, the standard deviation of the human data along component $i$, so eigen noise initially follows the spread of human postures. For \ours{}, we rescale $\sigma_q$ and $\sigma_e$ so that, summed over joints, the initial noise variance matches Joint-Space, split equally between the two.

\subsection{\ours{} for Reinforcement Learning}
\label{sec:reinforcement-learning}

Given an observation $o$, the policy network predicts $\mu_q=\pi_\theta(o)$, which we use as the joint-space command $a_q$. PPO~\citep{ppo} learns the policy parameters $\theta$ and the noise standard deviations $\sigma_q$ and $\sigma_e$, while $E$ remains fixed. Conditioned on the observation $o$, the \ours{} sampling rule in Eq.~(\ref{eq:formulations}) is equivalently written as
\begin{equation}
    q^\star\sim\mathcal{N}(\mu_q,\Sigma),\qquad
    \Sigma=\operatorname{diag}(\sigma_q^2)
    +E^\top\operatorname{diag}(\sigma_e^2)E.
    \label{eq:rl-distribution}
\end{equation}
The covariance $\Sigma$ adds the independent joint-noise covariance to the covariance of the projected eigen noise. PPO evaluates action log probabilities and entropy under this distribution. At deployment, we execute $\mu_q$ deterministically and discard $E$ and the exploration parameters.

\subsection{\ours{} for Trajectory Optimization}
\label{sec:method-integration}

In trajectory optimization, exploration refines a sequence of joint targets rather than learning a policy. In SPIDER~\citep{spider}, these sequences are optimized to track reference hand and object motion. \ours{} leaves SPIDER's joint-space trajectory representation and objective unchanged, and modifies only the perturbations used to sample candidate trajectories. At each iteration, the optimizer generates $N$ candidate sequences by perturbing the current joint target $q_t$ at each time $t$ with the \ours{} noise in Eq.~(\ref{eq:formulations}), using $q_t$ in place of $a_q$. It evaluates each candidate trajectory in simulation, keeps the lowest-cost 10\% of trajectories, indexed by $\mathcal{S}$, and updates each target by adding a weighted average of the selected perturbations,
\begin{equation}
    q_t\leftarrow q_t+\sum_{j\in\mathcal{S}}w_j\left[
    \sigma_q\odot\eta_{q,t}^{(j)}
    +E^\top\!\left(\sigma_e\odot\eta_{e,t}^{(j)}\right)
    \right].
    \label{eq:trajectory-update}
\end{equation}
Here, $\eta_{q,t}^{(j)}$ and $\eta_{e,t}^{(j)}$ are the Gaussian noise vectors of candidate $j$ at time $t$, scaled by SPIDER's noise schedule, while $\sigma_q$ and $\sigma_e$ remain fixed. The weights $w_j$ sum to one, with larger weights for lower-cost candidates.

\section{Experiments}
\label{sec:experiments}

\begin{figure}[t]
\figurelayout
\includegraphics[width=\linewidth,trim=0 9.7bp 0 3.4bp,clip]{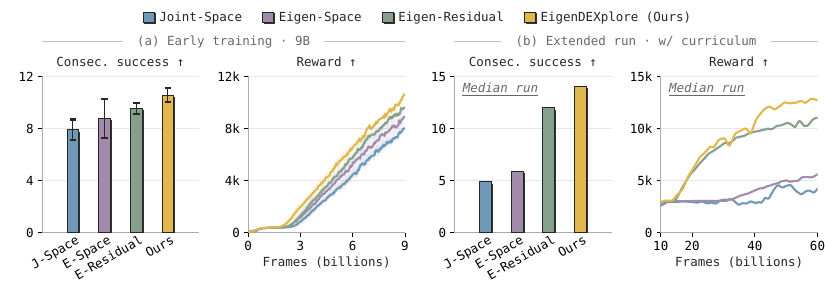}
\figurecaption{\textbf{SimToolReal.} \ours{} achieves the highest mean consecutive successes and reward in dexterous tool manipulation. (a) Fixed-budget performance at 9B frames. Error bars show 95\% CIs and shading shows $\pm$1 SE across five seeds. (b) Extended training with a success-tolerance curriculum using one median run per method.}
\label{fig:simtoolreal}
\end{figure}

\subsection{Action spaces and exploration}
\label{sec:action-space-comparison}

We compare all four formulations (Sec.~\ref{sec:coordinated-exploration}) in SimToolReal and DeXtreme over five seeds, reporting consecutive successes (number of object-pose goals reached).

\subsubsection{SimToolReal}
\label{sec:simtoolreal}

We adapt the training pipeline of SimToolReal~\citep{simtoolreal}, which trains a Sharpa hand-arm policy to manipulate procedurally generated tools toward target poses. Unlike the original, which narrows the Sharpa hand's abduction joint ranges, we use the full joint ranges for all methods. Following its original ablation budget, we first train for 9B frames with a fixed success tolerance of 6\,cm and $50^\circ$, emphasizing grasping and coarse reorientation. Eigen-Space and Eigen-Residual improve over Joint-Space, while \ours{} achieves the highest reward and a mean of 10.6 consecutive successes vs. 7.9 for Joint-Space (Fig.~\ref{fig:simtoolreal}a).

We then continue a single run per method, its median seed by consecutive successes at 9B frames, to 60 billion total frames with a curriculum that tightens the tolerance toward 1\,cm, $8^\circ$, requiring finer manipulation. We use one run per method because each extended run takes ${\sim}1$ week. Under these tighter requirements, Eigen-Space's early advantage fades, while Eigen-Residual and \ours{} sustain stronger learning, with \ours{} reaching 14.1 consecutive successes by the end of training, compared with 12.0 for Eigen-Residual and 4.9 for Joint-Space (Fig.~\ref{fig:simtoolreal}b).

\subsubsection{DeXtreme}
\label{sec:dextreme}

\begin{wrapfigure}[18]{r}{0.5\textwidth}  
\vspace*{-18pt}
\figurelayout
\includegraphics[width=\linewidth,trim=0 9.7bp 0 2.3bp,clip]{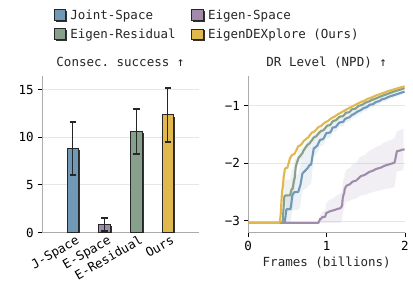}
\figurecaption{\textbf{DeXtreme.} \ours{} achieves the highest mean consecutive successes for in-hand reorientation.
\textbf{Left:} performance at 2B frames with DR Level fixed at $-0.5$. \textbf{Right:} DR Level during training. Error bars show 95\% CIs and shading shows $\pm$1 SE across five seeds.}
\label{fig:dextreme}
\end{wrapfigure}
DeXtreme~\citep{dextreme} trains an Allegro hand to repeatedly reorient a large cube toward randomly sampled target orientations. We train each variant for 2 billion environment frames with automatic domain randomization (ADR), which widens the ranges of observation noise, action delays, and physical properties as the policy meets a success threshold. The domain randomization (DR) level is measured in nats per dimension (NPD), the average log width of the randomized ranges. Higher values indicate further curriculum progression. Negative values correspond to widths below one.

Reorienting a large cube demands finger configurations that may be rare in human postures. Eigen-Space performs poorly, as it is confined to a 10-D subspace of Allegro's 16 joints. \ours{} achieves the highest mean consecutive successes under shared randomization (12.3 vs.\ 8.8 for Joint-Space) and progresses earlier through the DR curriculum (Fig.~\ref{fig:dextreme}).

Across both benchmarks, \ours{} improves over Joint-Space without changing the action representation, Eigen-Space struggles with more demanding manipulation, and Eigen-Residual trails \ours{} slightly, consistent with a cost to expanding the action space. Additional delta-action variants show similar trends and are reported in App.~\ref{sec:appendix-action-spaces}.

\subsection{DexMachina}
\label{sec:dexmachina}

\begin{figure}[t]
\figurelayout
\includegraphics[width=\linewidth,trim=0 0bp 0 3bp,clip]{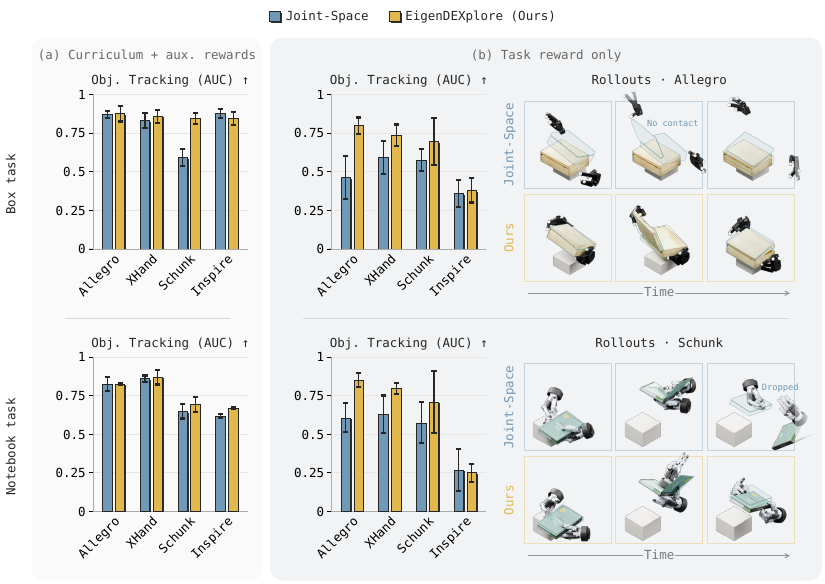}
\figurecaption{\textbf{DexMachina.} With curriculum and auxiliary rewards, \ours{} performs similarly to Joint-Space in most settings (a). However, with task reward alone, \ours{} substantially improves object tracking on higher-DoF Allegro and XHand (b, left). Representative task-reward-only rollouts (b, right). Bars show mean ADD-AUC with 95\% CIs across five seeds.}
\label{fig:dexmachina}
\end{figure}

DexMachina~\citep{dexmachina} trains bimanual policies to track object motions from ARCTIC demonstrations. To study how the benefit of \ours{} varies with training assistance and hand embodiment, we compare it with Joint-Space on box and notebook manipulation using Allegro (16 DoF), XHand (12), Schunk (9), and Inspire (6), with five seeds and about 2B frames per configuration. We report ADD-AUC, which summarizes object-tracking accuracy across error thresholds. We evaluate two levels of training assistance: the full setup, whose auxiliary motion and contact rewards and virtual-controller curriculum initially guide the object along the demonstration before withdrawing, and the object-tracking task reward alone.

With curriculum and auxiliary rewards, \ours{} performs similarly to Joint-Space in most settings, with a notable improvement for Schunk on the box task (0.59 to 0.84 ADD-AUC, Fig.~\ref{fig:dexmachina}a). This guidance already addresses much of the exploration difficulty, leaving less room for structured exploration to improve performance. With task reward alone, \ours{} achieves substantially higher object-tracking accuracy than Joint-Space, with the largest gains on the higher-DoF Allegro and XHand, for example from 0.46 to 0.80 ADD-AUC on the Allegro box task, and little change on Inspire (Fig.~\ref{fig:dexmachina}b). In the representative rollouts, \ours{} maintains stable contact while opening and closing both objects, whereas Joint-Space never touches the box with Allegro and drops the notebook mid-task with Schunk. The larger gains on higher-DoF hands are consistent with structured exploration helping most when coordinated actions must be discovered in a large search space without guidance. Lower-DoF hands such as Inspire are already highly coordinated, with three components capturing 90\% of the variance, leaving less room for improvement.

\subsection{SPIDER}
\label{sec:spider}

We use SPIDER~\citep{spider} to evaluate \ours{} in sampling-based trajectory optimization across hand embodiments. We optimize ten object-motion tracking tasks from OakInk2~\citep{oakink2} and compare Joint-Space with \ours{} on six hands: Sharpa (22 DoF), Wuji (20), Allegro (16), XHand (12), Schunk (9), and Inspire (6). Both methods use the same tasks, a cold start from the same fixed initial hand pose, and a fixed budget of 16 optimizer updates per commit, by which most runs have plateaued, with ten seeds per task. We report the retargeting cost at the end of optimization, where lower is better. This cost is SPIDER's optimization objective for tracking the reference hand and object trajectory.

\begin{figure}[t]
\figurelayout
\includegraphics[width=\linewidth,trim=0 0.2bp 0 2.3bp,clip]{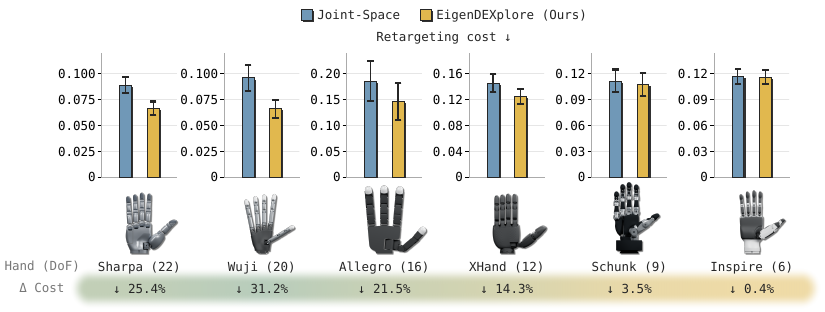}
\figurecaption{\textbf{SPIDER across hands.} 
\ours{} reduces mean retargeting cost across all hands, with the largest reductions on higher-DoF hands and the smallest on lower-DoF hands. $\Delta$ cost (\%) is relative to the Joint-Space baseline. Error bars show 95\% CIs over ten seed-wise task averages.
}
\label{fig:spider-hands}
\end{figure}

\ours{} reduces mean retargeting cost on every hand, and the reduction grows with DoF (Fig.~\ref{fig:spider-hands}). The largest reductions are on the high-DoF Wuji (31\%) and Sharpa (25\%), followed by Allegro (22\%) and XHand (14\%), while costs on Schunk and Inspire are nearly unchanged (4\% and 0.4\%), mirroring the trend in DexMachina (Sec.~\ref{sec:dexmachina}). Because SPIDER keeps noise scales fixed rather than learned (Sec.~\ref{sec:method-integration}), these gains come from the sampling distribution alone.

\subsection{Dataset and exploration analysis}
\label{sec:ablations}

\begin{figure}[t]
\figurelayout
\includegraphics[width=\linewidth,trim=0 9bp 0 5.9bp,clip]{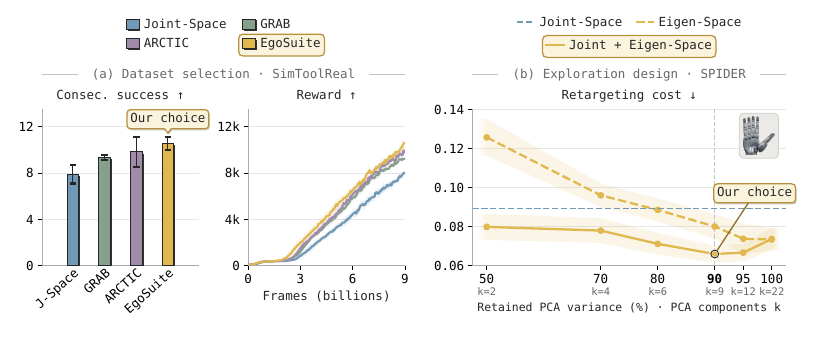}
\figurecaption{\textbf{Dataset and exploration analysis.} (a) Dataset selection: EgoSuite achieves the highest mean consecutive success and training reward in SimToolReal at 9B frames. (b) Exploration design: Sharpa SPIDER ablations vary the retained PCA variance (and thus the number of PCA components $k$) with and without joint-space noise. The selected 90\% setting with joint-space noise achieves the lowest mean retargeting cost. Shading shows $\pm1$ SE and error bars show 95\% CIs.}
\label{fig:dataset-analysis}
\end{figure}

We examine two design choices in \ours{}: the source dataset used to construct the human-motion prior and the exploration configuration.

For dataset selection, we compare priors fitted to GRAB~\citep{grab}, ARCTIC~\citep{arctic}, and EgoSuite~\citep{egosuite} in SimToolReal, holding the exploration configuration fixed and following the protocol in Sec.~\ref{sec:simtoolreal}. These datasets differ in scale and manipulation coverage: GRAB provides 1,334 motion-capture sequences involving 51 rigid objects, ARCTIC provides 339 sequences involving 11 articulated objects, and EgoSuite provides 100k hours of egocentric video spanning more than 15k tasks. All three priors improve over Joint-Space, and EgoSuite achieves the highest mean consecutive successes at 9B frames, 10.6 compared with 9.8 for ARCTIC, 9.4 for GRAB, and 7.9 for Joint-Space (Fig.~\ref{fig:dataset-analysis}a), supporting its use as the source dataset for \ours{}.

For exploration design, we fix the prior to EgoSuite and sweep the retained-variance threshold $\rho\in\{0.50,0.70,0.80,0.90,0.95,1.00\}$ on Sharpa, with and without independent joint-space noise, using the SPIDER setup in Sec.~\ref{sec:spider}. This tests how strongly exploration should be concentrated in the human-motion subspace and whether exploration outside that subspace improves optimization. With eigen noise alone, retargeting cost decreases as more components are retained, and at $\rho=0.50$, it is worse than Joint-Space (0.126 vs. 0.089), indicating that a narrow subspace restricts the search. Combining eigen noise with joint-space noise, as in \ours{}, lowers cost at every threshold below $\rho=1.00$, reaching a minimum of 0.066 at $\rho=0.90$, 25\% below Joint-Space (Fig.~\ref{fig:dataset-analysis}b). We therefore use this configuration in the dataset comparison and our main experiments.

These trends hold across GRAB, ARCTIC, and EgoSuite priors (App.~\ref{sec:appendix-retention}). We also find that random orthonormal directions with the same variance spectrum perform worse than Joint-Space, so the gains come from human coordination patterns rather than correlation alone (App.~\ref{sec:appendix-random}).

\subsection{Real-world evaluation}
\label{sec:realworld}

\begin{figure}[!t]
\figurelayout
\includegraphics[width=\linewidth,trim=0 0 0 0,clip]{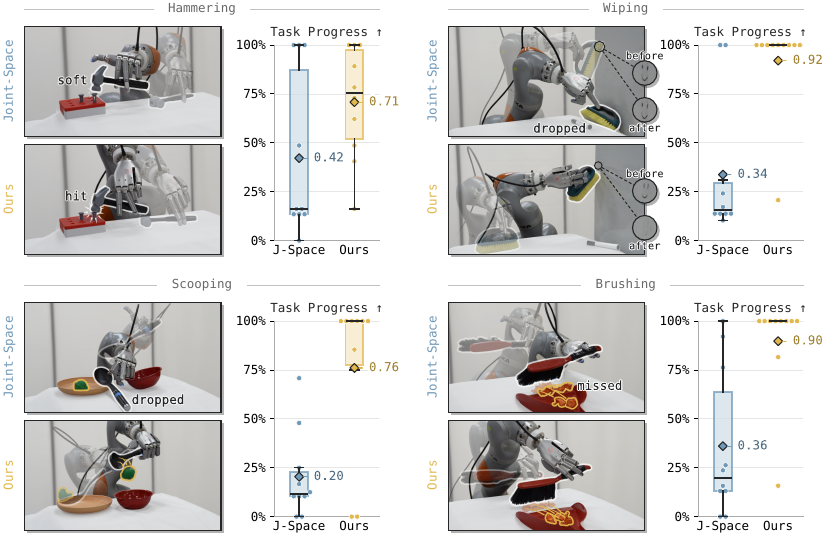}
\figurecaption{\textbf{Real-world SimToolReal.} \ours{} achieves higher task progress on tool-use. For each task, representative rollouts (left) and task progress, the fraction of waypoints reached (right). Boxes show quartiles/medians, diamonds show means, and dots show individual trials.}
\label{fig:realworld-simtoolreal}
\end{figure}

To test whether these improvements transfer to hardware, we compare \ours{} with the Joint-Space baseline on the SimToolReal and DeXtreme hardware setups.

\subsubsection{SimToolReal}
\label{sec:realworld-simtoolreal}

We evaluate the Joint-Space and \ours{} policies from the extended-training comparison in Sec.~\ref{sec:simtoolreal} on four tool-use tasks: hammering, wiping, scooping, and brushing (Fig.~\ref{fig:realworld-simtoolreal}). Both policies were trained for 60B frames, half of the 120B-frame budget used in the original SimToolReal training. The Joint-Space policy is trained in our adapted pipeline and is not the released SimToolReal checkpoint. We deploy them on the SimToolReal Sharpa hand-arm setup, using the same objects and trajectories as the original work~\citep{simtoolreal}. These trajectories consist of 6D object-pose waypoints extracted from human RGB-D videos using FoundationPose~\citep{foundationpose}. During execution, the policy advances to the next waypoint when the current goal pose is reached. We run ten trials per object-trajectory combination and report SimToolReal's Task Progress metric, the fraction of demonstrated waypoints reached. A trial ends at the final waypoint or on failure, such as losing the grasp or getting stuck.

\ours{} achieves higher mean task progress than Joint-Space on all four tasks, averaging 82\% across tasks compared with 33\% for Joint-Space (Fig.~\ref{fig:realworld-simtoolreal}). The Joint-Space policy typically pinches the tool between the pad of the middle finger and the backs of the index and ring fingers. It often requires several grasp attempts to lift the tool and struggles to reorient the tool in hand, instead relying on arm motion or dropping and regrasping. When the tool gets wedged deeply between the three fingers, the policy can complete the full trajectory, but the low friction on the backs of the fingers makes this grasp prone to slipping. In contrast, \ours{} learns a more conventional fingertip grasp using the pads of the thumb, index, and middle fingers, with support from the side of the ring finger. This grasp is more stable and enables in-hand reorientation, allowing the policy to reach successive goal poses more smoothly. This qualitative difference is consistent with the hypothesis that eigen-structured exploration makes coordinated, human-like grasp configurations easier to discover than independent joint-space exploration. Overall, these results show that the simulation advantage observed during extended training (Sec.~\ref{sec:simtoolreal}) transfers to hardware.

\Needspace{17\baselineskip}
\subsubsection{DeXtreme}
\label{sec:realworld-dextreme}

\begin{wrapfigure}[12]{r}{0.5\textwidth}  
\wrapfigureraise
\figurelayout
\includegraphics[width=\linewidth,trim=0 1bp 0 0.9bp,clip]{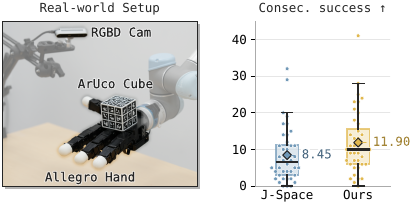}
\figurecaption{\textbf{Real-world DeXtreme.}
\ours{} increases mean consecutive successes from 8.45 to 11.90 over 40 trials per method. \textbf{Left:} hardware setup. \textbf{Right:} consecutive successes per trial. Boxes show quartiles/medians, diamonds show means, and dots show individual trials.
}
\label{fig:realworld-dextreme}
\end{wrapfigure}

Each method uses one policy trained with the 2-billion-frame budget described in Sec.~\ref{sec:dextreme}. A wrist-fixed Allegro hand repeatedly reorients a cube to successive target orientations (Fig.~\ref{fig:realworld-dextreme}). We estimate cube pose from ArUco markers using AprilCube~\citep{park2026aprilcube}. A trial ends when the cube is dropped or remains stuck in the same configuration for more than 80 seconds. Over 40 trials per method, \ours{} increases mean consecutive successes from 8.45 to 11.90 (41\% improvement) and the longest run from 32 to 41 (28\% improvement).

\section{Conclusion}
\label{sec:conclusion}

\ours{} adds perturbations along human-derived eigenvectors to independent joint-space noise while retaining joint-space actions. Across RL, trajectory optimization, six dexterous hands, and real-world deployment, it outperforms joint-space and learned action-space baselines, with the largest gains on higher-DoF hands and under less reward shaping. These results suggest that human-motion priors help mainly by structuring exploration rather than changing the action space.

\textbf{Limitations.} \ours{} uses a fixed linear basis derived from retargeted human motion, so coordination patterns absent from the data or nonlinear synergies must be explored through independent joint noise. Adapting the basis during training or using nonlinear priors could address this.
The prior depends on kinematic retargeting and may be less effective for hands that differ substantially from human morphology.
Improvements are also smaller for low-DoF hands. Finally, our noise is sampled independently at each timestep, and deriving temporally correlated exploration~\citep{gsde,pinknoise} from human motion is a natural extension.

\ificlrfinal
\section*{Acknowledgments}
We thank all members of IPRL and REALab at Stanford for helpful discussions and feedback. This work was supported in part by the National Science Foundation (NSF) under Grant Numbers 2342246, 2327974, 2143601, 2037101, and 2132519. Tyler Ga Wei Lum is supported by the Natural Sciences and Engineering Research Council of Canada (NSERC) under Award Number 526541680. The views and conclusions contained herein are those of the authors and should not be interpreted as necessarily representing the official policies, either expressed or implied, of the sponsors.
\fi

\section*{AI Use Statement}
We used generative AI tools to assist with implementing our method and benchmarking code, polishing the writing, generating figures programmatically from our results, and surveying related work to complement our own literature review. The authors verified all code and experiments, reviewed all AI-assisted content, and take full responsibility for this paper, including any text, claims, or artifacts produced with AI assistance.

\bibliographystyle{iclr2026_conference}
\bibliography{refs}

\clearpage

\appendix

\section{Additional Action Spaces}
\label{sec:appendix-action-spaces}

The formulations in Eq.~(\ref{eq:formulations}) command absolute joint targets. A common alternative integrates actions as increments onto the previous target~\citep{unicross,lucid}. We evaluate delta variants of Joint-Space and Eigen-Residual in this form,
\begin{equation}
    \begin{alignedat}{3}
    &\text{Joint-}\Delta &\quad&q^\star_t = q^\star_{t-1}+\actionterm{a_q}+\noiseterm{\sigma_q\odot\eta_q}
    &\quad&\text{action dim}=d\\
    &\text{Eigen-Residual-}\Delta &&q^\star_t = q^\star_{t-1}+\actionterm{a_q}+\noiseterm{\sigma_q\odot\eta_q}
    +E^\top\!\left(\actionterm{a_e}+\noiseterm{\sigma_e\odot\eta_e}\right)
    &&\text{action dim}=d+k
    \end{alignedat}
    \label{eq:delta-formulations}
\end{equation}
Here, $a_q$ and $a_e$ act as per-step increments, scaled to a fraction of each joint range and coefficient span, and $q^\star_t$ is clipped to the joint limits. Joint-$\Delta$ integrates independent joint increments as in~\citet{unicross}, and Eigen-Residual-$\Delta$ also integrates increments along the eigenbasis. LUCID~\citep{lucid} integrates eigen-grasp coefficients with per-joint residuals and is an instantiation of Eigen-Residual-$\Delta$.

\begin{figure}[h]  
\figurelayout
\includegraphics[width=\linewidth,trim=0 0bp 0 3.8bp,clip]{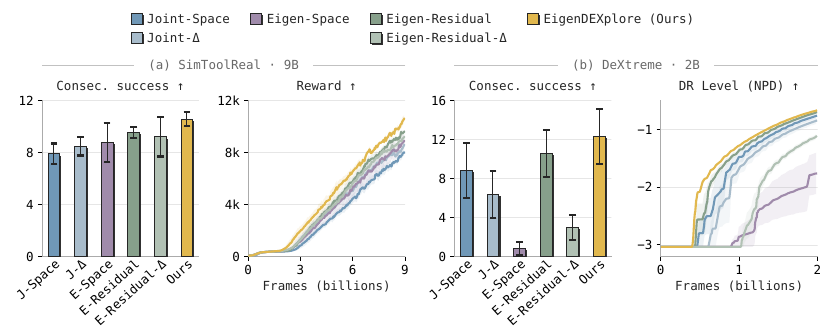}
\figurecaption{\textbf{Delta action spaces.} Delta variants of Joint-Space and Eigen-Residual perform comparably to their absolute counterparts in SimToolReal but fall behind them in DeXtreme. (a) SimToolReal at 9B frames, with consecutive successes (left) and reward (right). (b) DeXtreme at 2B frames, with consecutive successes at DR Level fixed at $-0.5$ (left) and DR Level during training (right). Error bars show 95\% CIs and shading shows $\pm$1 SE across five seeds.}
\label{fig:appendix-delta}
\end{figure}

We train both variants under the SimToolReal and DeXtreme protocols of Sec.~\ref{sec:action-space-comparison} (Fig.~\ref{fig:appendix-delta}). They perform comparably to their absolute counterparts in SimToolReal but fall well behind in DeXtreme. 

\section{Exploration Noise Analysis}
\label{sec:appendix-exploration}

\subsection{Source Dataset and Retained Variance}
\label{sec:appendix-retention}

\begin{figure}[t]
\figurelayout
\includegraphics[width=\linewidth,trim=0 13.6bp 0 3.0bp,clip]{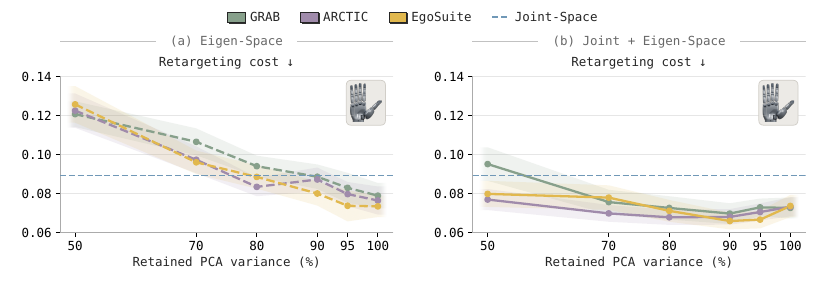}
\figurecaption{\textbf{Exploration design across datasets.} Priors fitted to GRAB, ARCTIC, and EgoSuite follow the same trends, so the benefit of combining eigen and joint-space noise does not depend on the source dataset. (a) With eigen noise alone, retargeting cost decreases as more components are retained. (b) Adding joint-space noise lowers cost at nearly every threshold. Dashed blue lines mark Joint-Space. Shading shows $\pm$1 SE.}
\label{fig:appendix-retention}
\end{figure}

We repeat the exploration-design sweep of Sec.~\ref{sec:ablations} with GRAB and ARCTIC priors in the same Sharpa SPIDER setup (Fig.~\ref{fig:appendix-retention}). All three priors follow the EgoSuite trends. Eigen noise alone improves as more human variance is retained, and adding joint-space noise lowers cost across thresholds, with the lowest cost between 80\% and 95\% retention for all three priors. This supports retaining 90\% of the PCA variance in \ours{}.

\subsection{Human-Derived versus Random Directions}
\label{sec:appendix-random}

\begin{figure}[t]
\figurelayout
\includegraphics[width=0.7\linewidth,trim=0 12.2bp 0 3.0bp,clip]{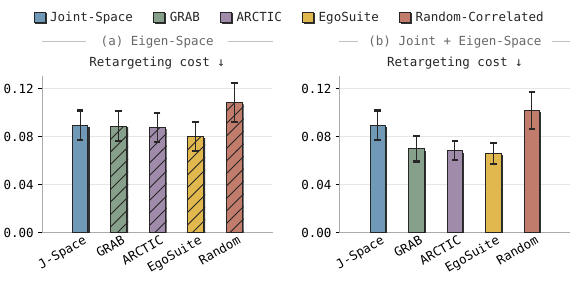}
\figurecaption{\textbf{Human-derived versus random directions.} The benefit of correlated exploration comes from the human-derived directions. Human-derived priors match or improve on Joint-Space, while random directions with the same eigenvalues are worse, both with eigen noise alone (a) and combined with joint-space noise (b). All eigen variants retain 90\% of the variance. Error bars show 95\% CIs over episodes.}
\label{fig:appendix-random}
\end{figure}

To test whether the benefit of \ours{} comes from the human-derived directions or from correlation alone, we introduce a Random-Correlated baseline. It replaces the eigenvectors in $E$ with random orthonormal directions but keeps the EgoSuite eigenvalues at 90\% retention as per-component variances. The eigenvalues are what correlate the noise, since equal variances along a full orthonormal basis would make it identical to joint-space noise. We evaluate it with eigen noise alone and combined with joint-space noise in the SPIDER setup of Sec.~\ref{sec:spider}. In both settings, random directions perform worse than joint-space noise, while the human-derived priors match or improve on it (Fig.~\ref{fig:appendix-random}). Correlation alone therefore does not explain the benefit of \ours{}, which relies on the coordination patterns in human motion.

\end{document}